\documentclass[10pt,twocolumn,letterpaper]{article}

\usepackage{3dv}
\usepackage{times}
\usepackage{epsfig}
\usepackage{graphicx}
\usepackage{amsmath}
\usepackage{amssymb}
\usepackage{algorithm}
\usepackage{algpseudocode}
\usepackage[normalem]{ulem}
\usepackage{multirow}
\usepackage{wrapfig}
\usepackage{authblk} 
\usepackage{url}

\threedvfinalcopy 
\def\threedvPaperID{50} 

\ifthreedvfinal\fi
\begin{document}

\title{Fast and Efficient Depth Map Estimation from Light Fields}

\author[1,2]{Yuriy Anisimov}
\author[1,2]{Didier Stricker}
\affil[1]{Department Augmented Vision, German Research Center for Artificial Intelligence (DFKI)}
\affil[2]{Department of Computer Science, University of Kaiserslautern}
\affil[ ]{\tt\small http://av.dfki.de}

\maketitle
\thispagestyle{empty}

\begin{abstract}
The paper presents an algorithm for depth map estimation from the light field images in relatively small amount of time, using only single thread on CPU. The proposed method improves existing principle of line fitting in 4-dimensional light field space. Line fitting is based on color values comparison using kernel density estimation. Our method utilizes result of Semi-Global Matching (SGM) with Census transform-based matching cost as a border initialization for line fitting. It provides a significant reduction of computations needed to find the best depth match. With the suggested evaluation metric we show that proposed method is applicable for efficient depth map estimation while preserving low computational time compared to others.
\end{abstract}
\section{Introduction} \label{introduction}
The term "light field" originated in a monograph of Gershun, translated in \cite{gershun1939light}, where it is formulated as the region of space, studied from the point of view of radiant energy transfer. Object of studies in this space is the rectilinear propagating ray with the radiant energy. This interpretation was then extended to the definition of plenoptic function, proposed by Adelson and Bergen in \cite{adelson1991plenoptic}, which describes the information from any of the observing positions in space and time, and can be interpreted as parametrization of every possible location, viewing direction, wavelength and point in time; or, alternatively, in form where angles of direction are replaced by spatial coordinates of image plane.

As stated in \cite{levoy1996light}, a light field can be determined as the radiance at a point in a given direction, and the plenoptic function may be reduced to 4-dimensional space. This leads us to the description of widely used representation of the ray in modern light fields --- two-plane parametrization: a plane of spatial coordinates $(u, v)$ stands for a 2-dimensional image in the light field, and angular plane $(s, t)$ represents the viewpoint. Each light ray can be denoted as intersection of these two planes. 

The proposed method considers a particular case of the light field, defined as a set of densely captured views of the scene. Shifting of the viewpoints for capturing can be performed by one or two moving directions, which for 2-dimensional images leads to 3-dimensional or 4-dimensional light field respectively.

Different ways of capturing light fields exist at present. One of the approaches is by using the so-called plenoptic camera \cite{lumsdaine2009focused} which consists of single image sensor with micro-lens array in front of it. Another way is a camera on moving stage, which allows shifting the device at equal length and capturing images of light field with same baseline. Similar capturing principle can be used with one- or two-dimensional arrays of multiple cameras. 
\begin{figure}
\begin{center}
\begin{tabular}{cccc}
\includegraphics[height=25mm]{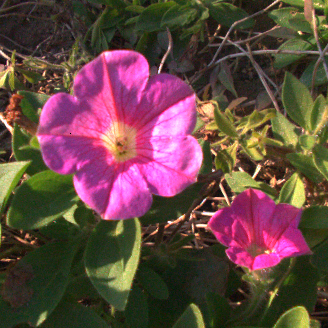} & \includegraphics[height=25mm]{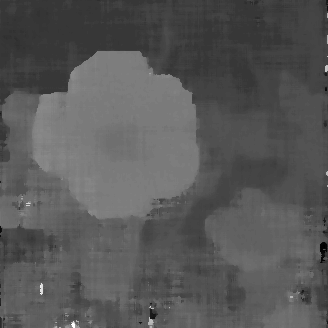} \\ \multicolumn{2}{c}{(a)}\\[6pt]
\includegraphics[height=25mm]{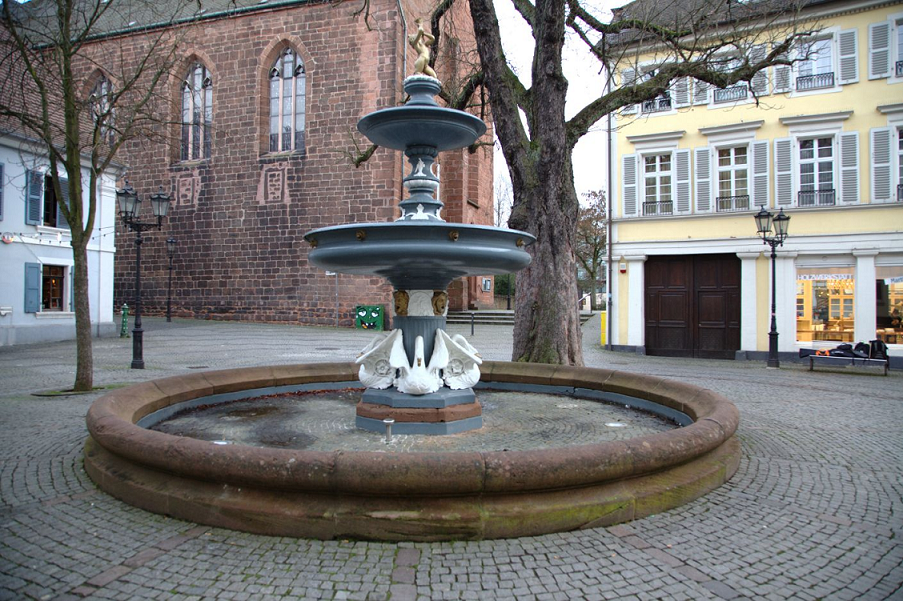} & \includegraphics[height=25mm]{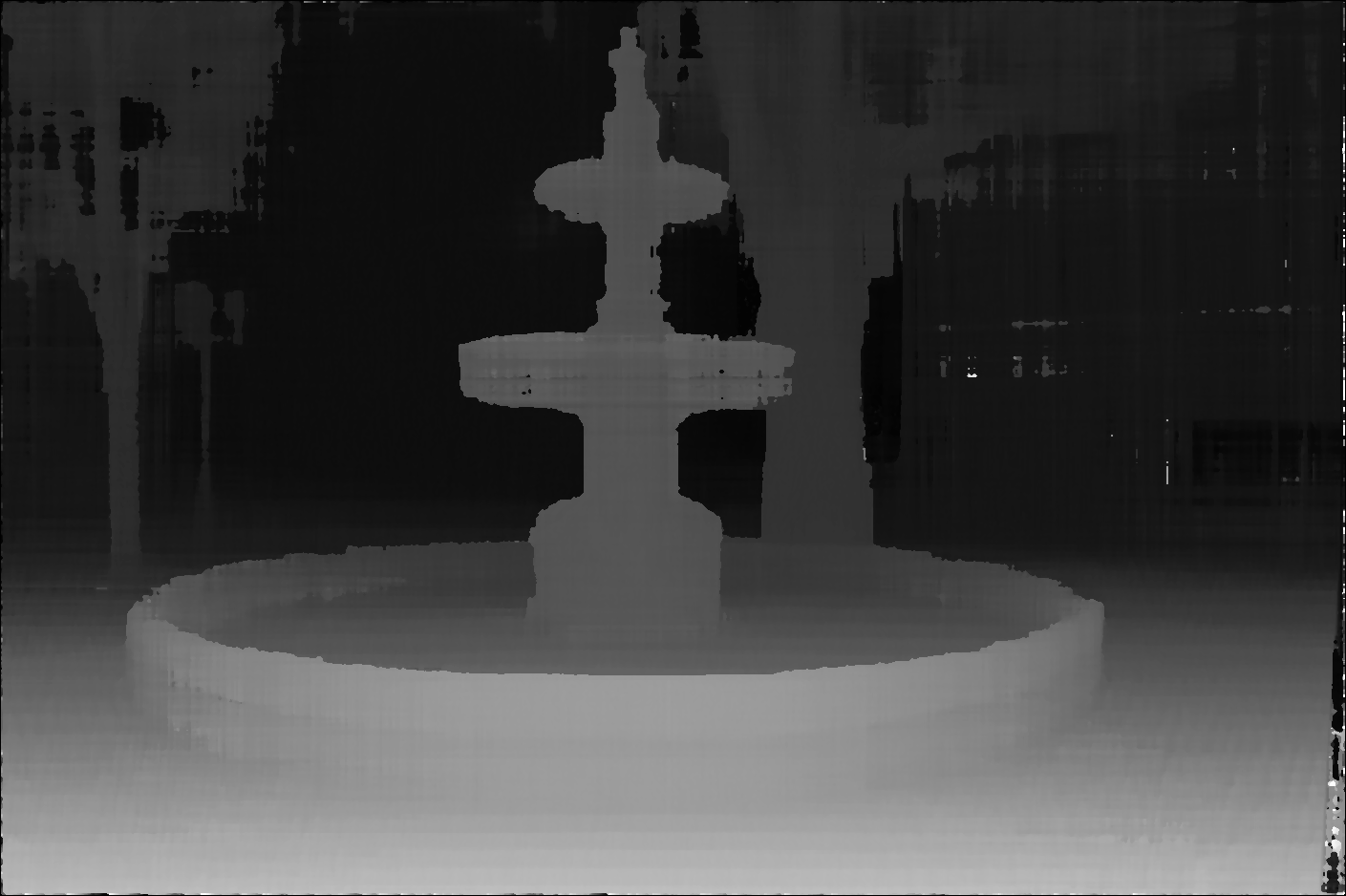} \\ \multicolumn{2}{c}{(b)}\\[6pt]
\end{tabular}
\caption{Results for real world scenes: (a) 4D light field, made from 7x7 images with resolution 328x328 from \cite{jeon2015accurate}, and a result of proposed method, (b) image from 3D light field, made from 29 images with resolution 1409x938, and a result of proposed method}
\label{fig:rw_scenes}
\end{center}
\end{figure}

Light field-based technologies and devices found their applications in industrial area (\eg optical inspection), in three-dimensional microscopy and in the cinema industry. All the light field applications require performing calculations related to scene depth value estimation. Large number of algorithms for such processing of light fields exist, and a short review of them is presented in Section \ref{related_work}. There is a trade-off between quality and runtime in the following algorithms: it takes a lot of time to process images with good results in the form of a depth map, or shorter amounts of processing time results in lower quality. Fast depth processing of light fields is not a trivial task, since big amount of data is involved and computations for the analysis methods are relatively complex. 

The proposed method aims to compute a dense depth map of acceptable quality from a given 4D light field in relatively small amount of time with the possibility of further optimization. The approach is based on the realization of line fitting concept from \cite{kim2013scene} with additional Semi-Global Matching (SGM)-based\cite{hirschmuller2005accurate} initialization. Results of SGM are good in terms of depth estimation for the whole image with low noise level, and the runtime of the algorithm is usually relatively small, but result for some image details (\eg object borders, fine structures) is not very precise. On the other hand, line fitting principle gives good results in terms of detail preservation, however, the runtime is relatively high and there is a large level of noise in homogeneous regions. Without SGM information, line fitting goes through all the possible depth hypotheses, which is computationally intensive; with the SGM, runtime is reduced proportionally to determined pixel level. Section \ref{algorithm_description} provides detailed description of proposed method. 

In Section \ref{experiments}, we show the output and evaluation of the algorithm, as well as faced limitations. Section \ref{conclusion} describes the conclusion and planned work, related to possible optimizations.

Our main contribution is the optimization of line fitting principle by utilizing results of SGM algorithm as bordering information with some optimizations.
Also, in Section \ref{experiments} we propose an evaluation metric related to number of correct pixels per second. With this metric it is showed that the performance of the algorithm is comparable to the state-of-the-art methods, while the time difference in most cases is significant.  

\section{Related work} \label{related_work}
\subsection{Light field analysis} \label{light_field_analysis}
One of the first studies related to depth map estimation from 3D light fields was proposed by Bolles \etal \cite{bolles1987epipolar}. It utilizes a structure derived from light field "slicing" and called "Epipolar-plane Image" (EPI). In the method detection of different features such as edges, peaks, and troughs is performed; results of the detection are used for line fitting in the EPI for structure estimation. Matou{\v{s}}ek \etal \cite{matouvsek2001accurate} propose a dynamic programming solution for correspondences detection in the EPI, based on finding lines with similar intensities and minimization of a cost function.

Kim \etal \cite{kim2013scene} propose a method for precise scene reconstruction from dense sequence of high-resolution images using fine-to-coarse strategy with further propagation. It based on the EPI, with the line fitting function algorithm tests several depth hypotheses and picks the one which leads to the highest color density. Computed values, which are checked for confidence, used as bounds for depth calculation after EPI downsampling. The algorithm in \cite{kim2013scene} is computationally intensive and can be efficiently realized only using GPU. However, the principle of line fitting, based on kernel density estimation, formed the basis for our method. Instead of downsampling, borders are obtained using SGM results.

Jeon \etal \cite{jeon2015accurate} compute a cost volume from the shifting of sub-aperture images with gradient- and color-based similarity measurement. Refinement of the depth map, obtained by a winner-takes-all strategy, is performed using graph cuts \cite{kolmogorov2002multi}. In \cite{wanner2012globally} and \cite{wanner2014variational}, Wanner and Goldluecke estimate a disparity map from light fields using EPI analysis with a structure tensor method, solving so-called "constrained labeling problem" with further variational regularization. Tao \etal \cite{tao2013depth} combines results of defocus and correspondence cue responses, calculated from the sheared EPI, to obtain the depth map. In \cite{wang2015occlusion} results of \cite{tao2013depth} are extended for the occlusion-handling case. Zhang \etal \cite{zhang2016robust} propose a spinning parallelogram operator for depth estimation on a EPI. 

Basha \etal \cite{basha2012structure} constructs dense volumetric representation of the three-dimensional space for estimating structure and motion using a multi-camera array. Because of the voxel representation of the scene, method can be considered as computationally-intensive. Neri \etal \cite{neri2015multi} presents multi-resolution method, based on local minimization of the maximum likelihood functional. The approach in \cite{yucer2016depth} utilizes the patch-based local gradient information for depth calculation with further propagation. Navarro and Buades \cite{navarro2017robust} propose a combination of two stereo non-dense methods, which in conjunction with interpolation gives a dense depth map. 
\subsection{Semi-global matching}
Since SGM was invented in \cite{hirschmuller2005accurate}, several works related to it were published. SGM, which is used for border initialization, is based on pipeline, described in \cite{haller2010real}. In proposed method, however, SGM uses cost aggregation method from the original paper \cite{hirschmuller2005accurate}, penalty parameter $P1$ is not excluded from the calculations. Sub-pixel interpolation is performed in the classic parabolic form. This approach was then improved by GPU implementation in \cite{haller2010gpu}. 

CPU implementation based on 5x5 Census cost matching, proposed by Gehrig and Rabe in \cite{gehrig2010real}. For reduction of memory consumption authors suggest to run SGM on subsampled images, on which one pixel is equal to the averaged value of four corresponding pixels in the original image. A pipeline for SGM with adaptive Census window is proposed by Loghman and Kim in \cite{loghman2013sgm}. An FPGA implementation of SGM is described in \cite{banz2010real}. Hermann and Klette in \cite{hermann2012iterative} propose the iterative approach for SGM with invention of semi-global distance maps as a cost function alternative, which reduces the amount of data needed to process.
\section{Algorithm description} \label{algorithm_description}
In this section, we explain the proposed algorithm. \ref{census_transform_based_matching_cost}-\ref{left_right_consistency_check} describe the steps for the line fitting borders initialization. These steps are performed for the first and last images of light field center row. Utilization of previous steps result in the form of a synthetic depth map described in \ref{synthetic_sgm}.  \ref{top_bottom_semi_global_matching} gives an explanation for improving border quality with additional computations using the first and last images in the light field center column. Confidence measurement for depth map values from \ref{left_right_consistency_check} with application to the synthetic depth map explained in \ref{confidence_measurement}. Additional edge filtration is provided in \ref{edges_exclusion}. Bordering information calculation is provided in subsection \ref{bordering_information} and the line fitting approach is described in \ref{line_fitting}. Aggregation for a final depth map presented in \ref{final_depth_map}.
\subsection{Census transform-based matching cost} \label{census_transform_based_matching_cost}
Census transform is a non-parametric transform, described in \cite{zabih1994non}. 
It is based on the comparison of radiance values between a pixel and its surroundings pixels within some window. To minimize the processing time while keeping capability of capturing whole image information, sparse Census window is used instead of densely filled one.
\begin{figure}
\begin{center}
\includegraphics[width=20mm]{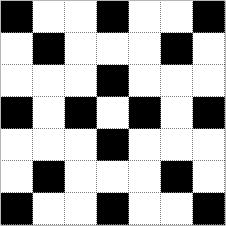}
\caption{A pattern for sparse Census processing}
\label{fig:census}
\end{center}
\end{figure}
The result of the transform is a bit string obtained as
\begin{align}
I_{c}(u,v)=\underset{[i,j]\in D}{\bigotimes}\xi(I(u,v),\,I(u+i,v+j)),
\end{align}
where $I_{c}$ is Census-transformed image, $I$ stands for the grayscaled image, $D$ is a set with coordinates of window elements used for transform, $\otimes$ stands for bitwise concatenation. Pixels relation is given by: 
\begin{align}
{\xi(p_1, p_2)} =
\begin{cases}
0, &p_1 \leqslant p_2\\
1,& p_1 > p_2
\end{cases}
.
\end{align}
In our approach, pixel sampling for cost function is performed in both possible shifting directions (forward and backward); it leads to a possibility of SGM calculations for such images, in which the background and foreground move to the different directions. 
Cost calculation between pixels in two Census-transformed images is determined as a Hamming distance \cite{hamming1950error} between pixels in both images
\begin{align}
C(u,v,d) = HD(I\textit{1}_{c}(u,v),I\textit{2}_{c}(u+d,v)),
\label{align:cost}
\end{align}
where $C$ -- structure with cost calculations result, $I\textit{1}_{c}$ and $I\textit{2}_{c}$ are first and second Census-transformed images, $d$ -- depth hypothesis, equal to pixel shift movement, which lies between maximum hypothesis level for moving in direction $d\textit{1}$ and $d2$ ($d\textit{1}_{max}$ and $d\textit{2}_{max}$ respectively). $HD$ stands for Hamming distance, which is calculated between two vectors $x_i$ and $x_j$ with same size $n$, as a quantity of elements with different values ($\oplus$ denotes exclusive disjunction)
\begin{align}
HD(x_i, x_j) = \sum\limits_{k=1}^n x_{ik} \oplus x_{jk}.
\end{align}
\subsection{Semi-global matching} \label{semi_global_matching}
Costs are aggregated using the principle from original SGM paper [16]. For each pixel $p=(u,v)$ and depth hypothesis $d$, after traversing in direction $r$ (determined as 2-dimensional vector with coordinate of pixel traversing $r$ = \{$\Delta$$u$, $\Delta$$v$\}), aggregated cost $L_r$ is  
\begin{align}
\begin{split}
&L_r(p,d) = C(p,d)+\\
&min\,(L_r(p-r, d),\\
&L_r(p-r, d-1) + P1,\\
&L_r(p-r, d+1) + P1,\\
&\underset{t}{min}\,L_r(p-r, t) + P2),
\end{split}
\end{align}
where $t$ lies between ${-d\textit{1}_{max}}$ and ${d\textit{2}_{max}}$.
Aggregated costs are summarized through all traversing directions:
\begin{align}
C_s(p,d) = \underset{r}{\sum}\,L_r(p,d).
\end{align}
From the cost summary, initial depth value is calculated using the winner-takes-all principle as
\begin{align}
D_{init}(p) =\underset{d}{\arg\min}\,C_s(p,d).
\end{align}
\subsection{Interpolation} \label{interpolation}
Refinement of the initial depth map is performed by parabolic interpolation of cost summary values 
\begin{align}
\begin{split}
&D_{sub}(p) = D_{init}(p)\; +\\ &\frac{C_s(p,d-1) - C_s(p,d+1)}{2*(2*C_s(p,d)-C_s(p,d-1)-C_s(p,d+1))}.
\end{split}
\end{align}
As a result of interpolation, depth values with sub-pixel accuracy are obtained. It gives a smoother outlook for the depth map without quality loss and allows us to calculate borders in \ref{bordering_information} more accurately. 
\subsection{Left-right consistency check}\label{left_right_consistency_check}
For occlusion filtering, steps \ref{census_transform_based_matching_cost}-\ref{interpolation} are calculated relative to left (left-right matching) and right (right-left matching) images, resulting in two sets of two depth maps \{$D_{Linit}$,$D_{Lsub}$\} and \{$D_{Rinit}$,$D_{Rsub}$\}.
Reliability of depth value in two depth maps for pixel $p$ is estimated through confidence measure
\begin{align}
{CMT_{LR}(p)} =
\begin{cases}
1,& |D_{L}(p) - D_{R}(p)| < \varphi\\
0,& \text{otherwise}
\end{cases},
\label{align:cmt}
\end{align}
where $\varphi$ stands for confidence threshold. 
\subsection{Synthetic depth map} \label{synthetic_sgm}
Previously obtained depth maps are used for construction of a new synthetic depth map, called $D_{syn}$. For that, interpolated depth maps need to be shifted to fit the position of the central view of the light field. Several methods can be used here for the translation distance determination: with known baseline and camera parameters translation can be calculated directly, in other cases information of phase correlation between images might be applicable.
For shifted depth maps $D_{Lsub_S}$ and $D_{Rsub_S}$, per-pixel averaging of the depth values is performed:
\begin{align}
D_{syn}(p)= (D_{Lsub_S}(p) + D_{Rsub_S}(p))/2.
\label{align:synthetic}
\end{align}
\subsection{Top-bottom Semi-global matching} \label{top_bottom_semi_global_matching}
For improving the quality of the result, similar calculations can be done for the first and last images in light field center column. (\ref{align:cost}) becomes 
\begin{align}
C(u,v,d) = HD(I\textit{1}_{c}(u,v),\,I\textit{2}_{c}(u,v+d)),
\end{align}
and steps, described in \ref{semi_global_matching}-\ref{left_right_consistency_check} are repeated with respect to changed moving direction (from top to bottom). 

As a result, we obtain sets of depth maps \{$D_{Tinit}$,$D_{Tsub}$\},\{$D_{Binit}$,$D_{Bsub}$\}, and confidence measurement $CMT_{TB}$. 
These depth maps can be used together with \{$D_{LsubS},D_{RsubS}$\} for the construction of a more accurate synthetic depth map. Shifting on depth maps (from \ref{synthetic_sgm}) need to be performed, and for the set of shifted depth maps \{$D_{LsubS},D_{RsubS},D_{TsubS},D_{BsubS}$\}, (\ref{align:synthetic}) becomes
\begin{align}
\begin{split}
&D_{syn}(p)=\\&(D_{Lsub_S}(p) + D_{Rsub_S}(p)\;+\\&(D_{Tsub_S}(p) + D_{Bsub_S}(p))/4 .
\label{align:synthetic_2}
\end{split}
\end{align}
\subsection{Confidence measurement} \label{confidence_measurement}
\begin{figure}
\centering
\begin{tabular}{ccc}
\includegraphics[height=35mm]{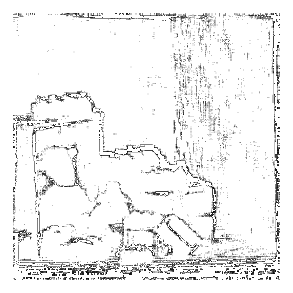} & \includegraphics[height=35mm]{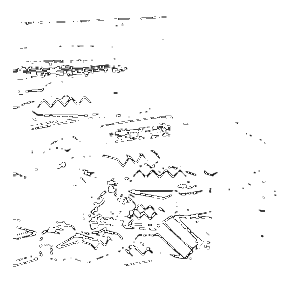}\\[6pt]
(a)&(b)
\end{tabular}
\caption{Intermediate image structures of the algorithm (with top-bottom SGM) for "dino" dataset from \cite{honauer2016benchmark}: (a) $CMT_{syn}$ in \ref{confidence_measurement}, (b) $E_{syn}$ in \ref{edges_exclusion}}
\end{figure}
Confidence measurement, previously mentioned in \ref{left_right_consistency_check}, is used for definition of a similar one for synthetic depth map $CMT_{syn}$. For left-right only matching, it is equal to $CMT_{syn}$; with included top-bottom SGM result is calculates as
\begin{align}
{CMT_{syn}(p)} =
\begin{cases}
1,& CMT_{LR}(p) = CMT_{TB}(p)\\
0,& \text{otherwise}
\end{cases}.
\label{align:cmt_syn}
\end{align}
\subsection{Edges exclusion} \label{edges_exclusion}
As mentioned before, SGM does not provide precise results on boundaries of the objects. This motivates us to exclude the bordering information for edges, obtained from center view of light field. For this, edges are calculated using the Sobel operator \cite{sobel19683x3} and smoothed using median filter; corresponding points are then stored in $E_{syn}$ structure. According to the experiments, this strategy works better in terms of precision of the final result than a scan using borders on edges from the SGM result. However, amount of data needed to be processed by the line fitting algorithm increases in this case, which affects the runtime.
\subsection{Bordering information} \label{bordering_information}
In order to calculate borders for line fitting concept, an intermediate structure $D_{brd}$ is created. Pixels, considered as unreliable in $CMT_{syn}$ and edges from $E_{syn}$, are marked in this structure to be calculated without border values, so full scan for every possible value will be done. For other values, normalization to line fitting coordinates is performed. Here, we introduce two parameters for the line fitting: depth window $DW$ and depth step $DS$. They determine, respectively, the range of search for the line to be fitted in light field and slope of the minimal depth hypothesis. Number of depth hypotheses for line fitting calculated as $N = DW/DS$.
From this, coefficient for normalization of bordering information is computed as
\begin{align}
k_{brd} = N / (d\textit{1}_{max} + d\textit{2}_{max}). 
\end{align}
With this data, values in $D_{brd}$ can be estimated through 
\begin{align}
D_{brd}(p) = (D_{syn}(p) + d\textit{1}_{max})k_{brd}
\end{align}
if $CMT_{syn}(p) = 1$ and $E_{syn}(p) = 0$;
and 0 otherwise. 
Using the $D_{brd}$ structure, low and high bordering values for each pixel are calculated with the algorithm 1 ($\lambda$ is a border penalty parameter) and stored respectively in $B_{L}$ and $B_{H}$ structures.

\subsection{Line fitting} \label{line_fitting}
As mentioned in \ref{light_field_analysis}, line fitting principle origins from \cite{kim2013scene}. We use the density estimation function to calculate depth score in areas bordered by previous steps. As a pivot image for calculations the center image of the light field is selected. It is denoted as $(\hat s=\lceil n/2 \rceil,\hat t=\lceil m/2 \rceil)$ in coordinate system of the light field, where $n$ and $m$ -- number of horizontal and vertical views in light field respectively. For each possible hypothesis of light field pixel of coordinates $(u,v)$ with respect to the pivot image density value is calculated as 
\begin{align}
\begin{split}
&S(u,v,d) ={\sum\limits_{s=1}^n\sum\limits_{t=1}^m}\\&{K(}{L(u+(\hat s - s)d,v+(\hat t - t)d,s,t)\;-}\\&{L(u,v,\hat s, \hat t))},
\label{align:S}
\end{split}
\end{align}
where $L$ is 4-dimensional light field, $s$ and $t$ -- horizontal and vertical position of image in light field.

Radiance similarity is verified using a optimized Parzen estimation method \cite{duda2012pattern} with a Epanechnikov kernel \cite{epanechnikov1969non}. For a given vector $x$:
\begin{align}
&{K(x)} =
\begin{cases}
1 - l,& l \leqslant 1 \\
0,& \text{otherwise}
\end{cases}\\
&l = \sum\limits_{i=1}^c x_i^2\upsilon\\
&\upsilon = 1 / h^2,
\label{align:band}
\end{align}
where $h$ stands for the bandwidth parameter, $c$ --- number of elements in vector. 

Unlike the original approach \cite{kim2013scene}, we use a simplified calculation of the kernel, \eg we avoid square root calculations, which reduces the computational time.
\begin{algorithm}
	\caption{Calculation of borders for line fitting}
	\begin{algorithmic}[1]
		\For{p = 1 to pixels} \do\\
				\If{$D_{brd}(p) = 0$}
            				 \State $B_{L}(p) = 0$
            				 \State $B_{H}(p) = N$
				\Else
					\If{$D_{brd}(p) - \lambda \geqslant 0$}
						\State $B_{L}(p) = D_{brd}(p) - \lambda$
					\Else
                 					\State $B_{L}(p) =0$
					\EndIf
            				\If {$D_{brd}(p) + \lambda \leqslant N$}
               					\State $B_{H}(p) = D_{brd}(p) + \lambda$
            				\Else
						\State $B_{H}(p) = N$
            				\EndIf
           			 \EndIf
		\EndFor
	\end{algorithmic}
\end{algorithm}
\subsection{Final depth map} \label{final_depth_map}
The final depth result is determined for the center light field view according to the highest value of the density sampling and saved in $D_{final}$:
\begin{align}
D_{final}(u,v) =\underset{d}{\arg\max}\,S(u,v).
\end{align}
Median filter is applied to final result to remove noise.

For memory efficiency purposes, values of S in (\ref{align:S}) are not stored during the processing time, after highest score estimation they are overwritten by the processing information for next pixels. 

\section{Experiments} \label{experiments}
In this section we provide the comparison of the proposed method with presented in Section \ref{related_work} state-of-the-art algorithms \cite{zhang2016robust,SAG17:cvpr,neri2015multi,si2016dense,johannsen2016sparse,wanner2012globally,jeon2015accurate,wang2015occlusion}. In tables and figures these algorithms are presented under acronyms SPO, OFSY, RM3DE, SC\_GC, EPI1, EPI2, LF and LF\_OCC respectively. Evaluation is carried out by the 4D Light Field Benchmark \cite{4DLFB} \cite{honauer2016benchmark}. 
\subsection {Datasets}
\begin{figure*}
\begin{center}
\begin{tabular}{cccccc}
\includegraphics[height=25mm]{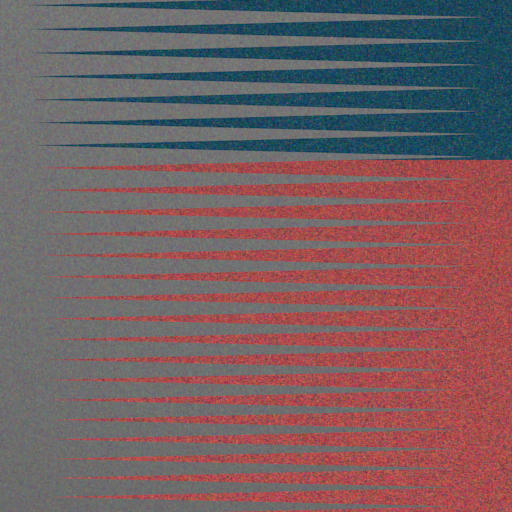} & \includegraphics[height=25mm]{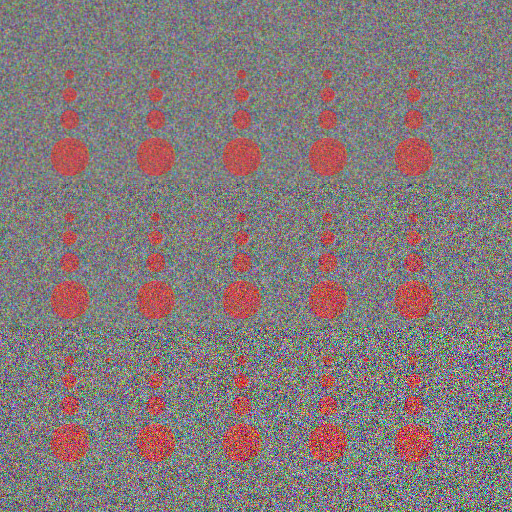} & \includegraphics[height=25mm]{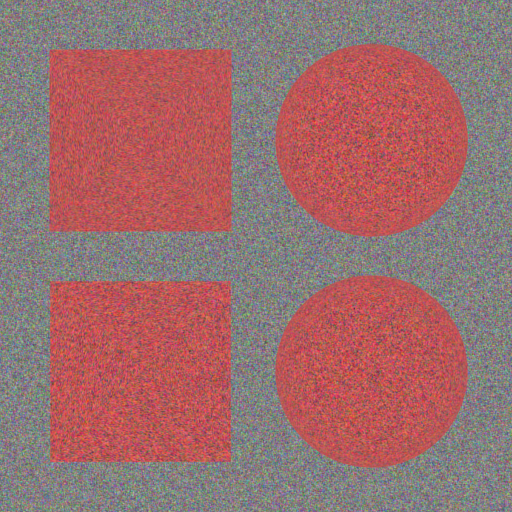} & \includegraphics[height=25mm]{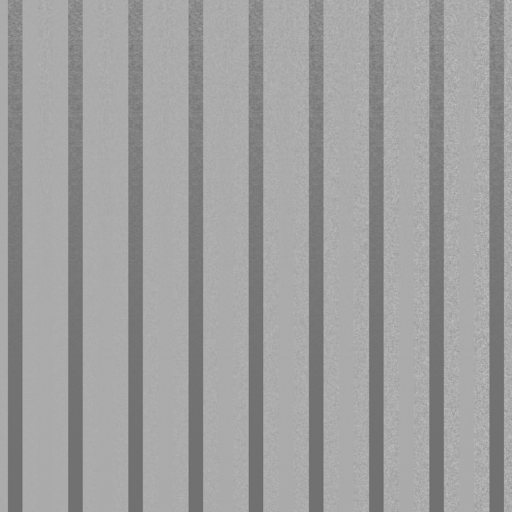} & \includegraphics[height=25mm]{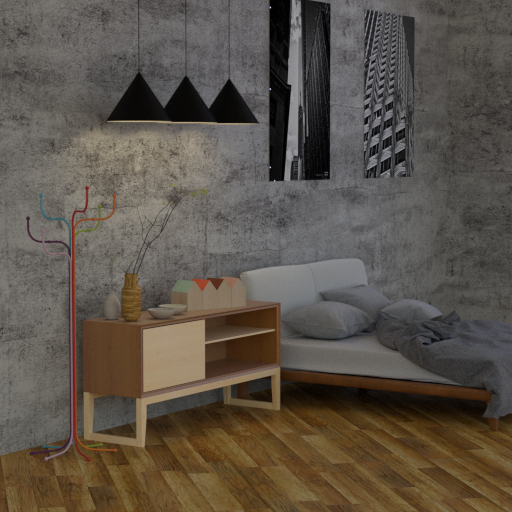} & \includegraphics[height=25mm]{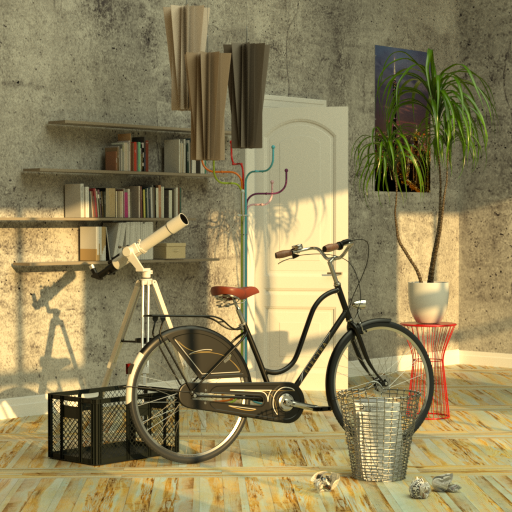} \\
(a)&(b)&(c)&(d)&(e)&(f) \\
\includegraphics[height=25mm]{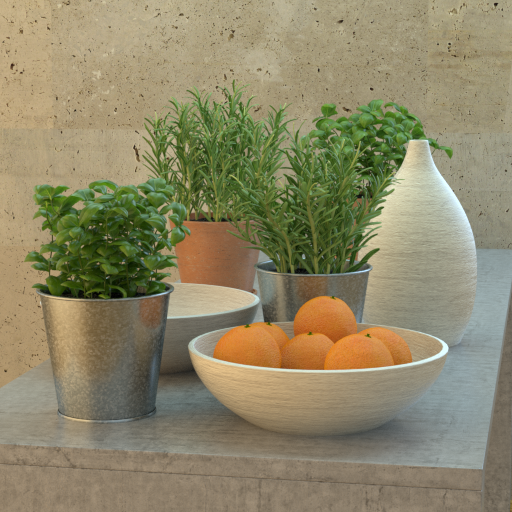} & \includegraphics[height=25mm]{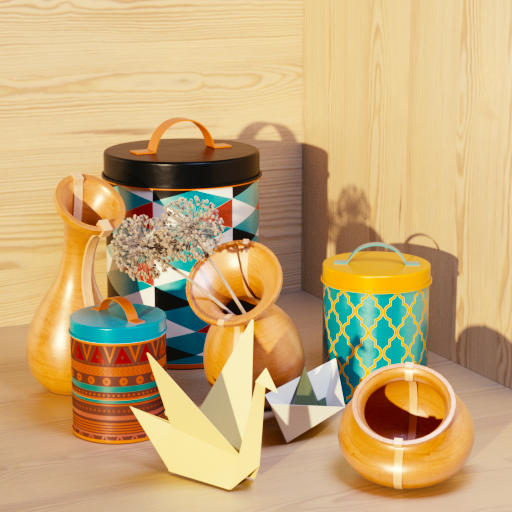} & \includegraphics[height=25mm]{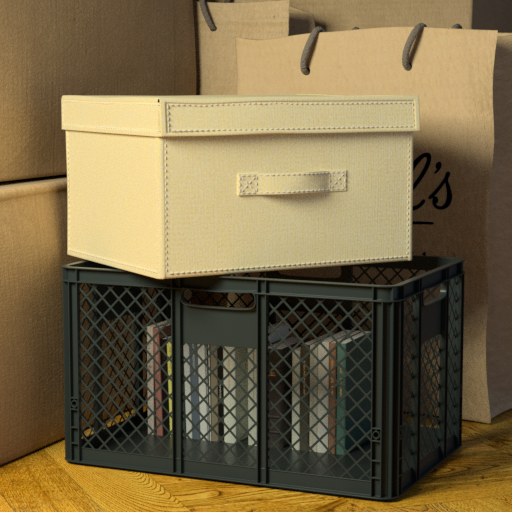} & \includegraphics[height=25mm]{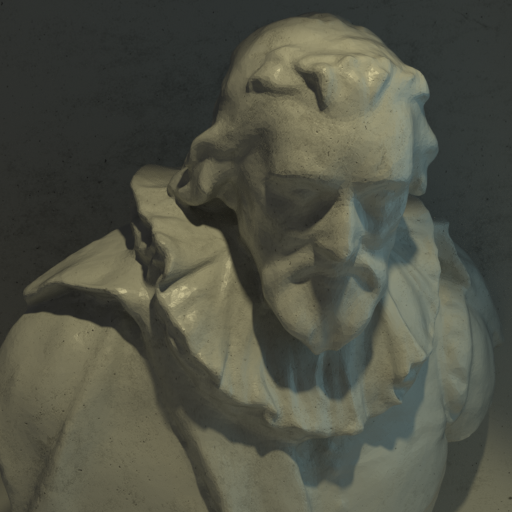} & \includegraphics[height=25mm]{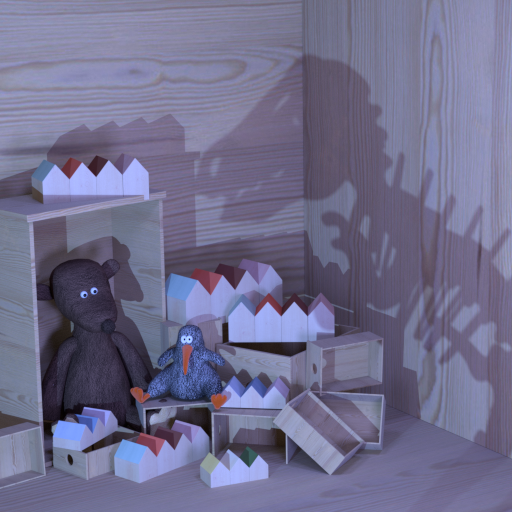} & \includegraphics[height=25mm]{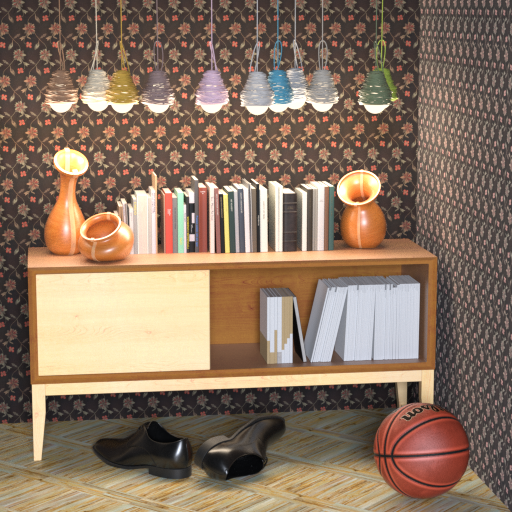} \\
(g)&(h)&(i)&(j)&(k)&(l)
\end{tabular}
\caption{Center images of light fields from \cite{honauer2016benchmark}: (a) "backgammon", (b) "dots", (c) "pyramids", (d) "stripes", (e) "bedroom", (f) "bicycle", (g) "herbs", (h) "origami", (i) "boxes", (j) "cotton", (k) "dino", (l) "sideboard"}
\end{center}
\end{figure*}
We use the light field images, provided by Honauer \etal \cite{honauer2016benchmark} through 4D Light Field Benchmark. 12 synthetic scenes are provided for the main evaluation; each scene is represented by the 9x9 light field, composed from 8-bit RGB images with resolution of 512x512 pixels. Datasets are grouped in three categories: "training" for evaluation and parameters adjustment, "stratified" with special challenging cases, and "test" for "blind" verification. Camera settings and disparity ranges provided for every light field, high resolution disparity and depth maps are provided only for "training" and "stratified" datasets. In this section we present image result comparison for "dino" and "cotton" datasets; results for other datasets can be found at \cite{4DLFB} under the \textit{BSL} acronym, and in supplementary materials.
\subsection {Metrics} \label{metrics}
Benchmark provides several metrics for result evaluation. Together with some general measurements, like Mean Squared Error, algorithms can be evaluated in terms of some photorealistic terms, \eg surface smoothness. Main criteria for evaluation of our method is the estimation of the percentage of errors in algorithm result, formulated as the $BadPix$ metric in mentioned benchmark, and the running time of the algorithm. $BadPix$ stands for the percentage of pixels in which absolute difference of result and ground truth bigger than $T$, where $T$ set to 0.07. Corresponding formulas and description can be found in \cite{honauer2016benchmark}. 
For purposes of interpreting our result in terms of stated in Section \ref{introduction} contributions, we propose a metric $M$. This metric stands for percentage of correctly computed pixels per second, formulated as
\begin{align}
M = \frac{100\% - BadPix}{Runtime}\Big(\frac{\%}{sec.}\Big),
\label{align:metric}
\end{align}
We provide an average and median result out of metrics mentioned above in tables, result for all datasets separately, together with other metrics, can be found at \cite{4DLFB}. 
\subsection{Algorithm settings} \label{algorithm_settings}
Mentioned in Section \ref{algorithm_description} parameters were adjusted for the best evaluation result in $BadPix$ metric and the runtime. Algorithm parameters stayed fixed independently of scene parameters except the disparity range. Penalty parameters for the SGM $P1$ and $P2$ were set to 21 and 45. Number of possible disparities (pixel shifting in both possible directions) is adjusted accordingly to the data, provided in configuration files for each of the scene. For the Census transform, used in SGM, different patterns of the aggregation window have been evaluated, for the experiments we use 7x7 pattern (Fig. \ref{fig:census}). Another options for sparse Census window are listed in \cite{loghman2013sgm}. 

Edge exclusion was not performed, and SGM is calculated for the left-right image pair only. These adjustments related to runtime optimization of the algorithm.

The range for line fitting is set corresponding to disparity ranges for each of the datasets, and sampling line step is set as $(1/(N - 1))\tau$, where N corresponds to number of images in one light field dimension ($N = 9$ in our case), and $\tau$ stands for step coefficient which is set to $1/7$. The kernel size of the median filter for final filtration is 3. The confidence threshold $\varphi$ in (\ref{align:cmt}) has been set to 3 and border penalty $\lambda$ in Alg.1 -- to 2. Bandwidth parameter $h$ in (\ref{align:band}) is set to 0.02.

\begin{figure*}
\begin{center}
\setlength{\tabcolsep}{1pt}
\begin{tabular}{cccccc}
\includegraphics[width=30mm]{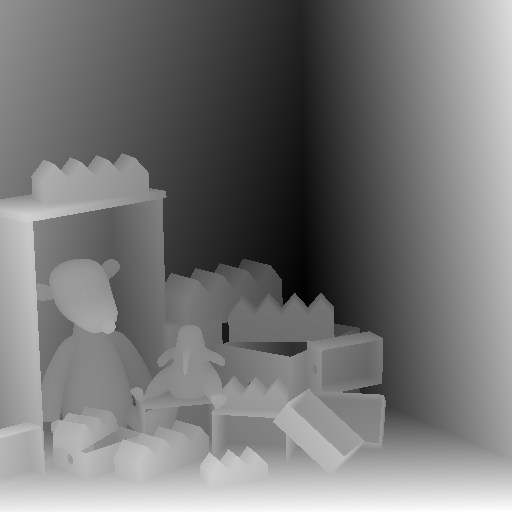} & \includegraphics[width=30mm]{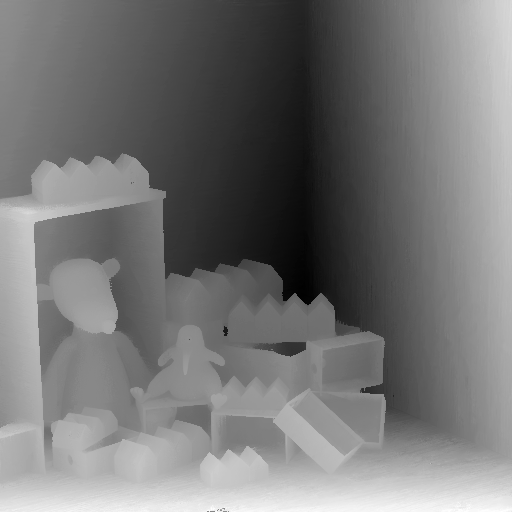} & \includegraphics[width=30mm]{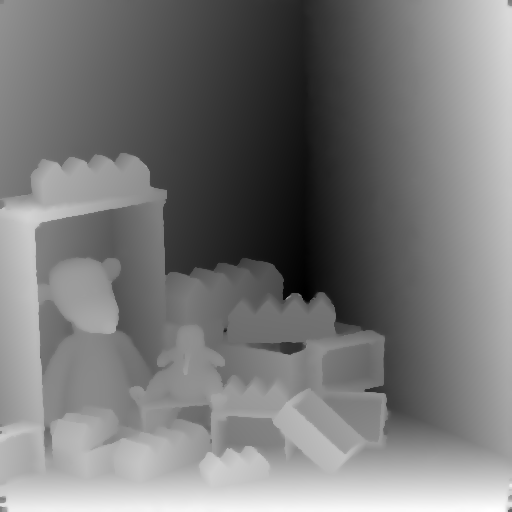} & \includegraphics[width=30mm]{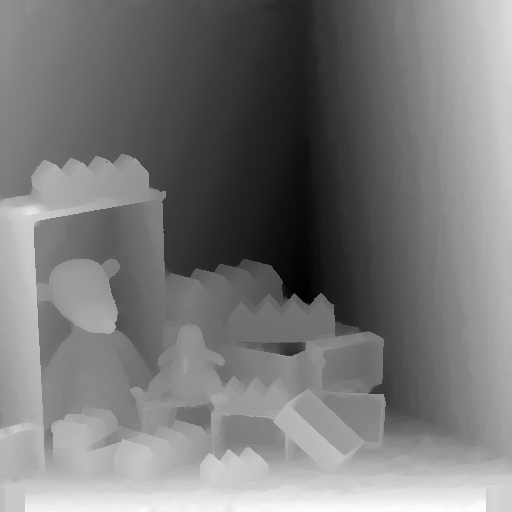} & \includegraphics[width=30mm]{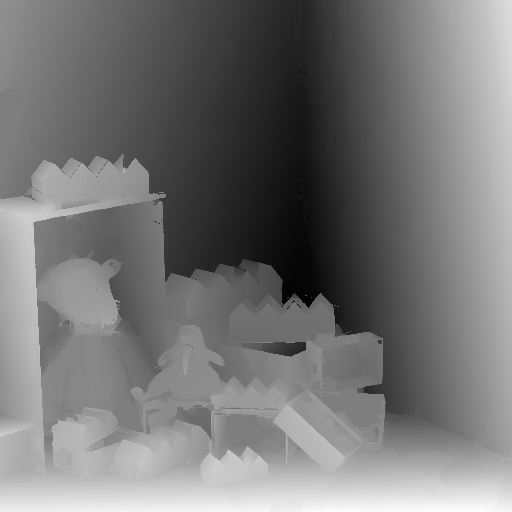} \\ (a) & (b) & (c) & (d) & (e) \\
\includegraphics[width=30mm]{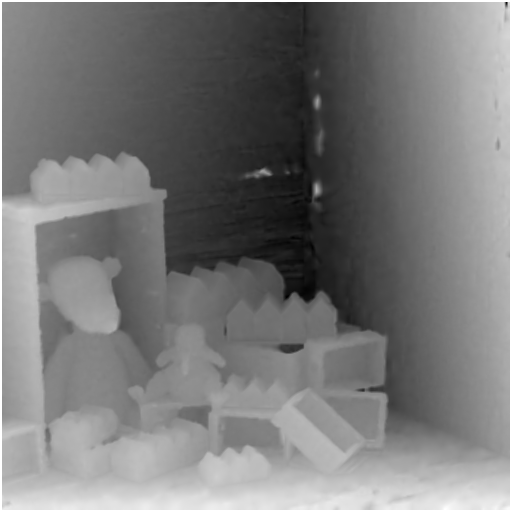} & \includegraphics[width=30mm]{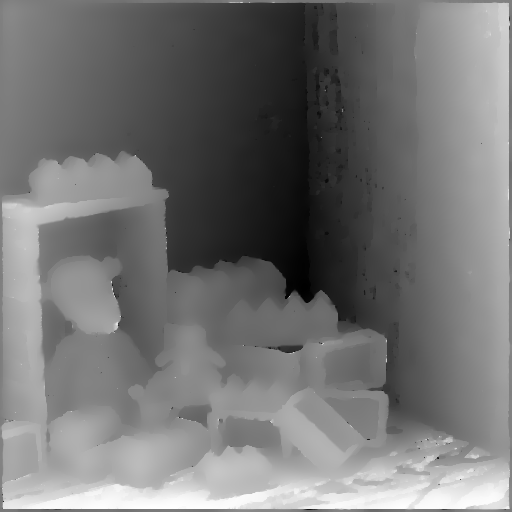} & \includegraphics[width=30mm]{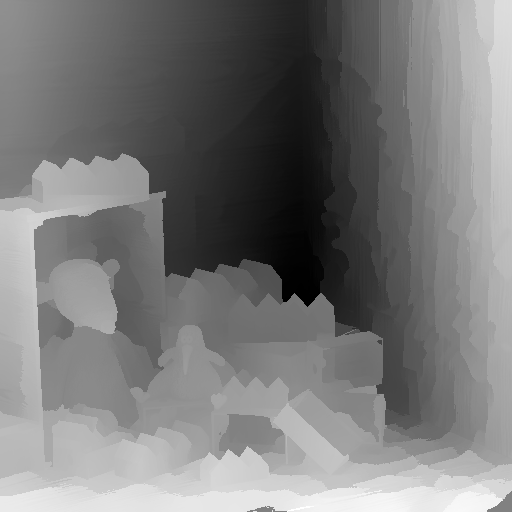} & \includegraphics[width=30mm]{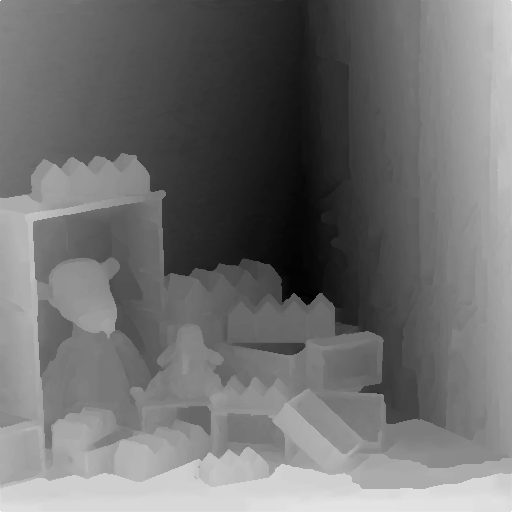} & \includegraphics[width=30mm]{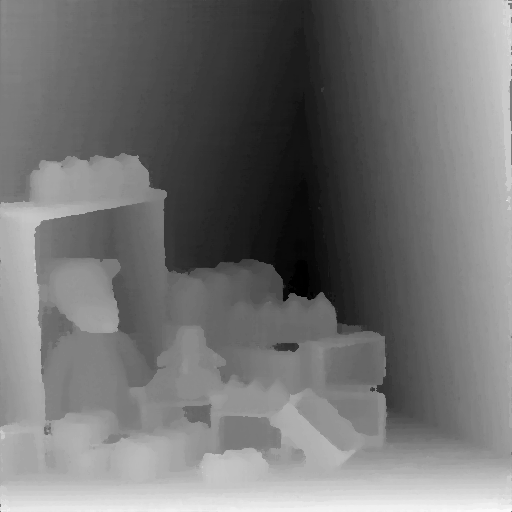} \\ (f) & (g) & (h) & (i) & (j) \\
\end{tabular}
\caption{Results for "dino" dataset from \cite{honauer2016benchmark}. (a) Ground truth, (b) SPO \cite{zhang2016robust}, OFSY \cite{SAG17:cvpr}, (d) RM3DE \cite{neri2015multi}, (e) SC\_GC \cite{si2016dense}, (f) EPI1 \cite{johannsen2016sparse}, (g) EPI2 \cite{wanner2012globally}, (h) LF \cite{jeon2015accurate}, (i) LF\_OCC \cite{wang2015occlusion}, (j) proposed method}
\label{fig:dino}
\end{center}
\end{figure*}
\subsection {Visualization}
\begin{figure*}
\begin{center}
\setlength{\tabcolsep}{1pt}
\begin{tabular}{cccccc}
\includegraphics[width=30mm]{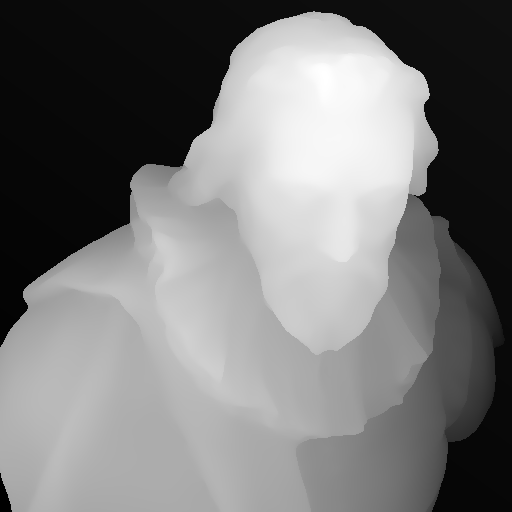} & \includegraphics[width=30mm]{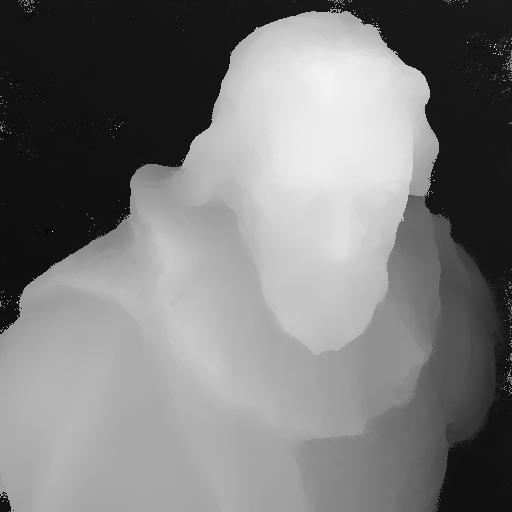} & \includegraphics[width=30mm]{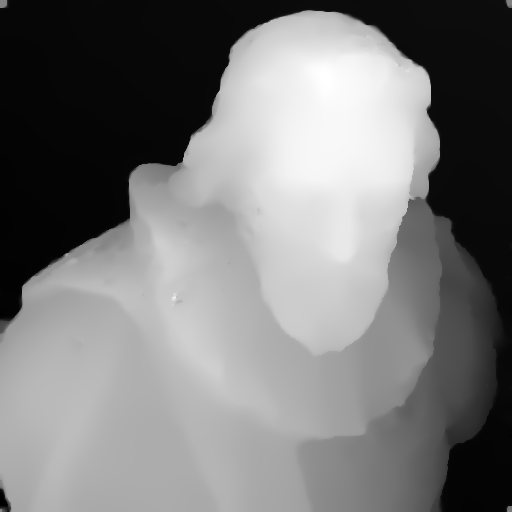} & \includegraphics[width=30mm]{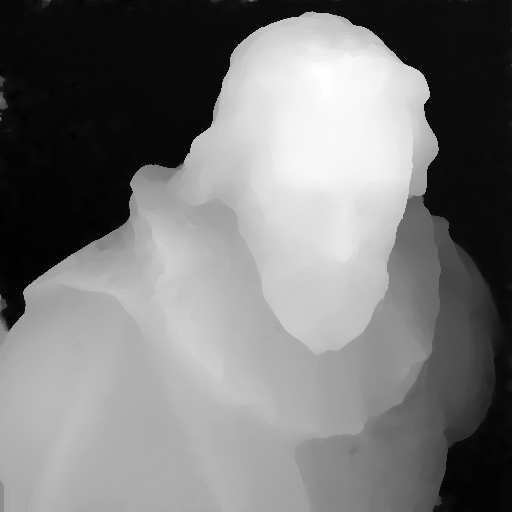} & \includegraphics[width=30mm]{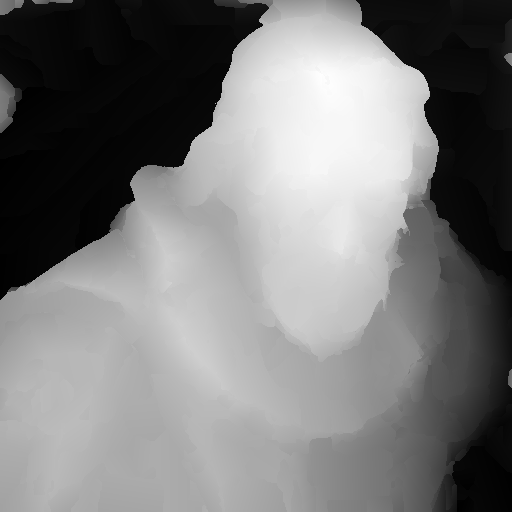} \\ (a) & (b) & (c) & (d) & (e) \\
\includegraphics[width=30mm]{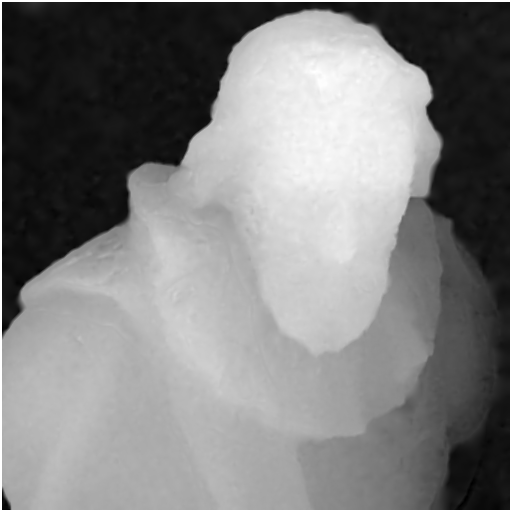} & \includegraphics[width=30mm]{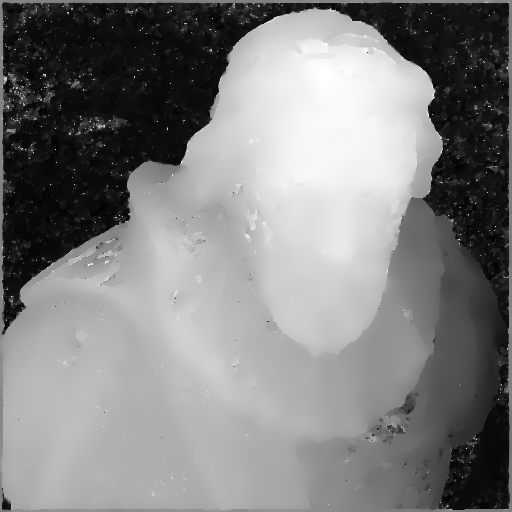} & \includegraphics[width=30mm]{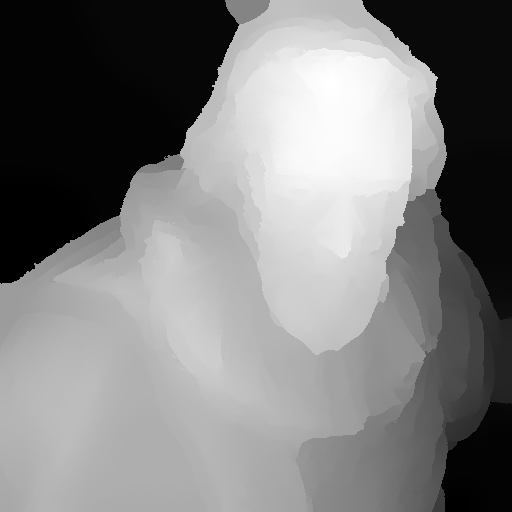} & \includegraphics[width=30mm]{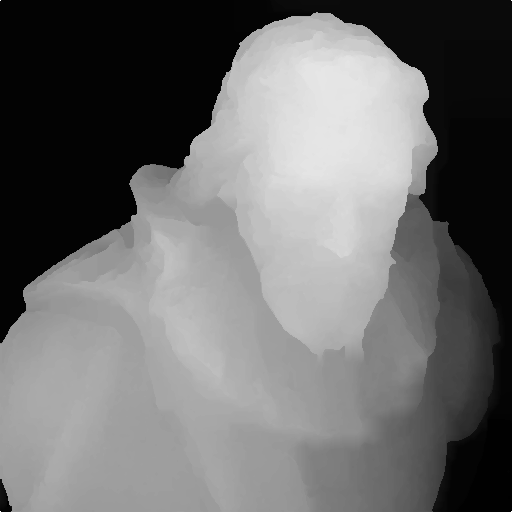} & \includegraphics[width=30mm]{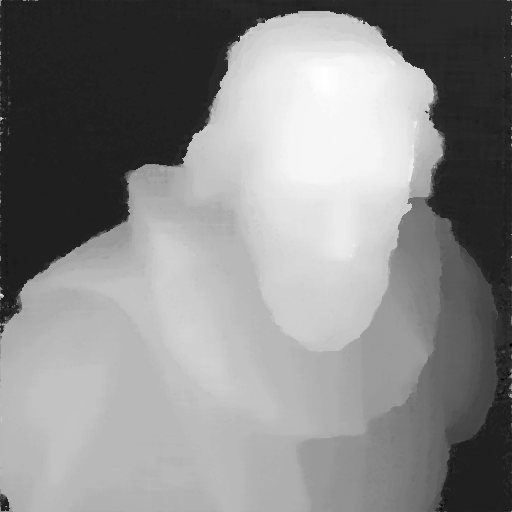} \\ (f) & (g) & (h) & (i) & (j) \\
\end{tabular}
\caption{Results for "cotton" dataset from \cite{honauer2016benchmark}. (a) Ground truth, (b) SPO \cite{zhang2016robust}, OFSY \cite{SAG17:cvpr}, (d) RM3DE \cite{neri2015multi}, (e) SC\_GC \cite{si2016dense}, (f) EPI1 \cite{johannsen2016sparse}, (g) EPI2 \cite{wanner2012globally}, (h) LF \cite{jeon2015accurate}, (i) LF\_OCC \cite{wang2015occlusion}, (j) proposed method}
\label{fig:cotton}
\end{center}
\end{figure*}

Fig. \ref{fig:dino} and \ref{fig:cotton} illustrate the depth estimation result on the "dino" and "cotton" light field datasets from benchmark.  
For "dino" dataset result faces some problems with sharpness of some fine structures; also, "step" effect on the wall can be noticed. Borders of some object are not accurate, and the same issue appears in the result of "cotton" dataset (noise on the head contour). In general, result of the proposed algorithm processing, compare to others, for these two datasets is on the average in terms of subjected level of quality. 

\begin{table}[]
\centering
\begin{center}
\begin{tabular}{|l|l|l|l|l|}
\hline
    & SPO \cite{zhang2016robust} & OFSY \cite{SAG17:cvpr} & RM3DE \cite{neri2015multi} \\ \hline
Median & 8.779          & 11.329  & \textbf{7.992} \\ \hline
Average & \textbf{8.466} & 12.036  & 10.216         \\ \hline\hline
    & SC\_GC \cite{si2016dense} & EPI1 \cite{johannsen2016sparse} & EPI2 \cite{wanner2012globally} \\ \hline
Median & 10.206         & 22.891  & 22.942         \\ \hline
Average & 14.299         & 24.324  & 22.651         \\ \hline\hline
    & LF \cite{jeon2015accurate} & LF\_OCC \cite{wang2015occlusion} & proposed \\ \hline
Median & 16.146         & 18.451  & 13.409         \\ \hline
Average & 16.193         & 17.579  & 12.743         \\ \hline
\end{tabular}
\end{center}
\caption{The percentage of pixels in which absolute difference of result and ground truth larger than threshold $T$ ($BadPix$) on 4D Light Field Benchmark \cite{honauer2016benchmark}. Here $T = 0.07$}
\label{table:badpix}
\end{table}

\begin{table}[]
\centering
\begin{center}
\begin{tabular}{|l|l|l|l|}
\hline
    & SPO \cite{zhang2016robust} & OFSY \cite{SAG17:cvpr} & RM3DE \cite{neri2015multi} \\ \hline
Median & 2111.500 & 198.299  & 45.149         \\ \hline
Average & 2115.417 & 200.282  & 47.434         \\ \hline\hline
    & SC\_GC \cite{si2016dense} & EPI1 \cite{johannsen2016sparse} & EPI2 \cite{wanner2012globally} \\ \hline
Median & 2052.190 & 85.045   & 8.789          \\ \hline
Average & 2056.344 & 88.194   & 8.406          \\ \hline\hline
    & LF \cite{jeon2015accurate} & LF\_OCC \cite{wang2015occlusion} & proposed \\ \hline
Median & 994.311  & 10614.54 & \textbf{5.149} \\ \hline
Average & 1009.756 & 10508.47 & \textbf{5.962} \\ \hline
\end{tabular}
\end{center}
\caption{Runtime in seconds on 4D Light Field Benchmark \cite{honauer2016benchmark}}
\label{table:runtime}
\end{table}

\subsection{Results}
Table \ref{table:badpix} shows the comparison of algorithm by \textit{BadPix} metric, described in subsection \ref{metrics}. Result of the proposed method is on the average position compare to others. 

Table \ref{table:runtime} represents the runtime of algorithms, reported by authors of evaluated methods in 4D Light Field Benchmark. According to this result, runtime of our method is better than in most of the state-of-the-art algorithms. In some scenes, SGM results were eliminated for a large number of pixels, and line fitting without bordering information for these pixels affected the runtime (datasets "bicycle", "herbs", "boxes" and "sideboard"). This occurs because of the amount of fine structures in the scene. Application of a smaller window for Census transform in SGM seems to be a solution in terms of accuracy of the borders for the fine structures; however, it reduces the quality of the whole image, hence results of these tests are not provided.
Some of the algorithms in the comparison were run on GPU architecture in contrast to our single thread algorithm. Parallelization for our method is available because of the nature of the algorithm. Parallel versions of SGM are covered in different papers \cite{banz2010real,haller2010gpu}, and for line fitting scan can be performed line-by-line in parallel, since no result of calculations on different lines is used.

Proposed algorithm outperforms the majority of algorithms in the benchmark in terms of percentage of correctly calculated pixels in the image per second. Results of the comparison with the proposed in (\ref{align:metric}) metric are presented in Table \ref{table:M}. For 12 images benchmark our result is the best in 10 cases; average and median value of the metric is the top-of-the-line.

\subsection{Real world scenes}
Fig. \ref{fig:rw_scenes} shows results for two real world scenes. First scene is represented as a 3D light field and contains 29 images with resolution of 1409x938 pixels. It was acquired using the moving stage with a camera on it. The baseline between each image is about 5 mm. For this dataset SGM penalty parameters $P1$ and $P2$ were set to 71 and 105 respectively. 

Second scene provided by Jeon \etal in \cite{jeon2015accurate} and obtained by lenslet-based light field camera. The 4D light field contains of 7x7 images with resolution of 328x328 pixels. Algorithm uses same parameters from \ref{algorithm_settings}, except step coefficient $\tau$, which was set to $1/5$. 

Visually the result is fine; however, some challenges, \eg small and fine structures, are noticeable. Processing of these scenes took 4.34 sec. and 411 ms respectively. We do not provide the comparison with other metrics from \ref{metrics}, since no ground truth is available for these images.

\subsection{Environment}
\begin{table}[]
\centering
\begin{center}
\begin{tabular}{|l|l|l|l|}
\hline
    & SPO \cite{zhang2016robust} & OFSY \cite{SAG17:cvpr} & RM3DE \cite{neri2015multi} \\ \hline
Median & 0.044  & 0.458   & 1.961           \\ \hline
Average & 0.043  & 0.465   & 1.952           \\ \hline\hline
    & SC\_GC \cite{si2016dense} & EPI1 \cite{johannsen2016sparse} & EPI2 \cite{wanner2012globally} \\ \hline
Median & 0.043  & 0.911   & 9.054           \\ \hline
Average & 0.042  & 0.867   & 9.310           \\ \hline\hline
    & LF \cite{jeon2015accurate} & LF\_OCC \cite{wang2015occlusion} & proposed \\ \hline
Median & 0.082  & 0.008   & \textbf{18.392} \\ \hline
Average & 0.083  & 0.009   & \textbf{22.247} \\ \hline
\end{tabular}
\end{center}
\caption{Percentage of correctly calculated pixels per second (\ref{align:metric}) on 4D Light Field Benchmark \cite{honauer2016benchmark}}
\label{table:M}
\end{table}
Execution of the proposed method was performed on CPU E3-1245 V2 @ 3.40 GHz, forced to work in a single thread. The proposed algorithm is implemented in C and compiled in Arch Linux using GCC v.7.1.1 with /O3 option. 

\subsection{Limitations}
During the experiments, several disadvantages of our approach have surfaced. Estimated depth maps are noisy in discontinuities area, for some images a "step" effect of depth change is preserved (datasets "bedroom", "boxes", "dino"); in the regions with random noise pattern, acceptable depth values are not calculated (dataset "dots"). Algorithm shows the average (and for a part of cases --- relatively bad) result in terms of evaluation of the proposed in \cite{honauer2016benchmark} photorealistic metrics. Quality-related optimizations need to be done for avoiding these problems. 

Also, subjective sharpness level of some objects is related to the selected configuration for the mentioned in \ref{synthetic_sgm} $D_{syn}$ structure. It drops if we use only left-right SGM result. Conjunction with top-bottom SGM (\ref{align:cmt_syn}) increases the object sharpness, but it requires more time for SGM calculations. Line fitting also suffers from that, since more pixels are marked as unreliable with further full scan for them.
\section{Conclusion} \label{conclusion}
In this paper, we presented an algorithm for depth estimation from light field images, which combines stereo matching and line fitting approaches. We verified our algorithm with the different metrics, and the result of evaluation showed us that the proposed method is achieving a comparable to state-of-the-art depth map result. The method shows one of the best runtime and outperforms most of the state-of-the-art algorithms with the proposed metric (\ref{align:metric}). For future work, we plan to add runtime-related modifications (parallelization, SIMD-instructions), use an adaptive Census window for the special cases and also involve the gradient information for matching and additional confidence measurements. 
\section*{Ackowledgments} \label{ackowledgments}
This work has been partially funded by the BMBF project DAKARA (13N14318). The authors are grateful to  Vladislav Golyanik, Kiran Varanasi and Jonathan Wray for the provided help.

{\small
\bibliographystyle{ieee}
\bibliography{egbib}
}
\end{document}